\documentclass[conference]{IEEEtran}
\IEEEoverridecommandlockouts

\usepackage{cite}
\usepackage{amsmath,amssymb,amsfonts}
\usepackage{algorithmic}
\usepackage{graphicx}
\usepackage{textcomp}
\usepackage{xcolor}
\usepackage{url}
\def\BibTeX{{\rm B\kern-.05em{\sc i\kern-.025em b}\kern-.08em
    T\kern-.1667em\lower.7ex\hbox{E}\kern-.125emX}}
\begin{document}

\title{Auditing Training Data in Domain-adapted LLMs: LoRA-MINT}

\author{

\IEEEauthorblockN{
Gonzalo Mancera\IEEEauthorrefmark{1},
Daniel DeAlcala\IEEEauthorrefmark{1},
Aythami Morales\IEEEauthorrefmark{1}\IEEEauthorrefmark{3}
Julian Fierrez\IEEEauthorrefmark{1},
Ruben Tolosana\IEEEauthorrefmark{1},
Francisco Jurado\IEEEauthorrefmark{2}
}

\IEEEauthorblockA{\IEEEauthorrefmark{1}BiometricsAI, Universidad Autonoma de Madrid, Spain}

\IEEEauthorblockA{\IEEEauthorrefmark{2}Computer Engineering Department, Universidad Autonoma de Madrid, Spain}

\IEEEauthorblockA{\IEEEauthorrefmark{3}Department of Mathematics, Universidad de Las Palmas de Gran Canaria, Spain}

}

\maketitle

\begin{abstract}
We present LoRA-MINT, a new methodology for Membership Inference Test (MINT) applied to recent Large Language Models (LLMs) fine-tuned for specific Natural Language Processing (NLP) tasks through Low-Rank Adaptation (LoRA). The primary goal is to assess whether individual samples were part of the training data of these adapted models, providing a useful auditing tool for the management of intellectual property and sensitive data. Our analysis explores the relationship between model perplexity and membership status, providing a systematic framework for estimating data exposure in fine-tuned LLMs. We conducted experiments on four models and three benchmark datasets, obtaining precision values in determining if given data were used for training ranging from 0.77 to 0.92, which outperform state-of-the-art baselines and demonstrate the robustness and generality of the proposed method. In general, our findings underscore the potential of LoRA-MINT as an effective and scalable framework for auditing LLMs, improving transparency, and fostering the ethical and responsible deployment of AI and NLP technologies. For the sake of concreteness and current relevance, our discussion and experiments are centered on LoRA-adjusted LLMs, but note that most of the presented methodology is easily applicable for auditing training data given any other technique for adapting LLMs or, more generally, any other domain-adapted AI models.
\end{abstract}

\section{Introduction}
In recent years, Natural Language Processing (NLP) technologies have achieved remarkable adoption in diverse domains — including healthcare \cite{locke2021natural}, law \cite{miguel26graph}, customer service \cite{olujimi2023nlp}, finance \cite{du2025natural}, human resources \cite{pena25llms}, and education \cite{iri26lak}. These systems have transformed the way textual information is processed, analyzed, and interpreted, thanks to advances in Machine Learning (ML) that enable the efficient handling of massive volumes of unstructured data. However, the accelerated integration of Artificial Intelligence (AI) and NLP solutions has also introduced complex ethical, social, and legal challenges \cite{sharma2025ethical,2023_SNCS_Human-Centric_Pena}, particularly in terms of privacy \cite{2017_Access_HEmultiDTW_Marta,2022_Access_DP-CL_Ahmad,omid2025digital,busch26book}, data ownership \cite{muldoon2025data}, and accountability \cite{dealcala2025active}.

In response to these emerging concerns, the European Union enacted the AI Act in June 2024\footnote{\url{https://artificialintelligenceact.eu}}, establishing one of the most comprehensive regulatory frameworks for AI to date. The legislation aims to ensure that AI technologies operate transparently, maintain fundamental rights, and remain auditable throughout their entire lifecycle. It explicitly promotes the creation of robust evaluation and monitoring mechanisms to detect potential misuse and guaranty regulatory compliance. Similar initiatives have been launched in other regions \cite{USA}, highlighting a global movement towards responsible AI governance, data traceability, and model accountability.

Within this evolving regulatory landscape, data privacy has become a central concern in the design and deployment of AI systems. As models grow in scale and capability, they are increasingly trained or adapted to use sensitive datasets that may include personal or private information \cite{chen2025survey}. Understanding how much of this information is implicitly memorized and potentially retrievable from trained models has therefore become essential for both ethical and legal compliance.

Within modern AI systems, Large Language Models (LLMs) have emerged as foundational technologies for state-of-the-art NLP \cite{zhao2023survey}. These models have evolved from general-purpose text generators into sophisticated systems capable of reasoning and decision support in professional domains — such as healthcare \cite{ goyal2024healai}, education \cite{iri26lak}, and finance \cite{wang2025financial}. Their remarkable capabilities stem from large-scale pre-training \cite{ achiam2023gpt}, during which they learn syntactic, semantic, and contextual patterns from massive text corpora. 

The fine-tuning process allows a general-purpose pretrained LLM to specialize in a particular task or domain by adjusting on smaller, curated datasets \cite{PENA-Layout}. However, since LLM architectures have grown in the billions of parameters, full fine-tuning has become computationally expensive and environmentally unsustainable \cite{patterson2021carbon}. This has led to the rise of Parameter-Efficient Fine-Tuning (PEFT) techniques, which aim to adapt LLMs using far fewer trainable parameters while maintaining strong performance.

Among these approaches, Low-Rank Adaptation (LoRA) \cite{hu2022lora} has gained exceptional traction due to its simplicity, scalability, and minimal resource requirements. Rather than updating every weight in an LLM, LoRA inserts small low-rank matrices into its linear layers, enabling targeted adaptation with only a fraction of the parameters. This dramatically reduces computational costs while preserving performance quality. Recent advances, such as quantized LoRA (QLoRA) \cite{dettmers2023qlora}, further reduce hardware demands, enabling efficient fine-tuning of billion-scale LLMs on commodity GPUs. 

However, while the accessibility of LoRA-based fine-tuning has democratized LLM customization, it has simultaneously intensified concerns about data exposure and privacy leakage \cite{carlini2021extracting}. Fine-tuning datasets — especially those from domains such as medicine, finance, or law — may contain personally identifiable or confidential information \cite{mancera2025pba}. Given the known tendency of LLMs to memorize parts of their training data, the discussion here is about extracting training data from LLMs. This discussion is particularly salient in the context of Membership Inference Attacks (MIAs) \cite{shokri2017membership}, where adversaries attempt to determine whether a particular data instance was used during model training by analyzing the model output. MIAs pose a direct challenge to the privacy of training data, and while some previous studies have examined these attacks in generic ML settings, few studies have specifically analyzed their behavior in fine-tuned LLMs, especially in those models adapted through LoRA.

\begin{figure}[t]
    \centering
    \includegraphics[width=0.98\linewidth]{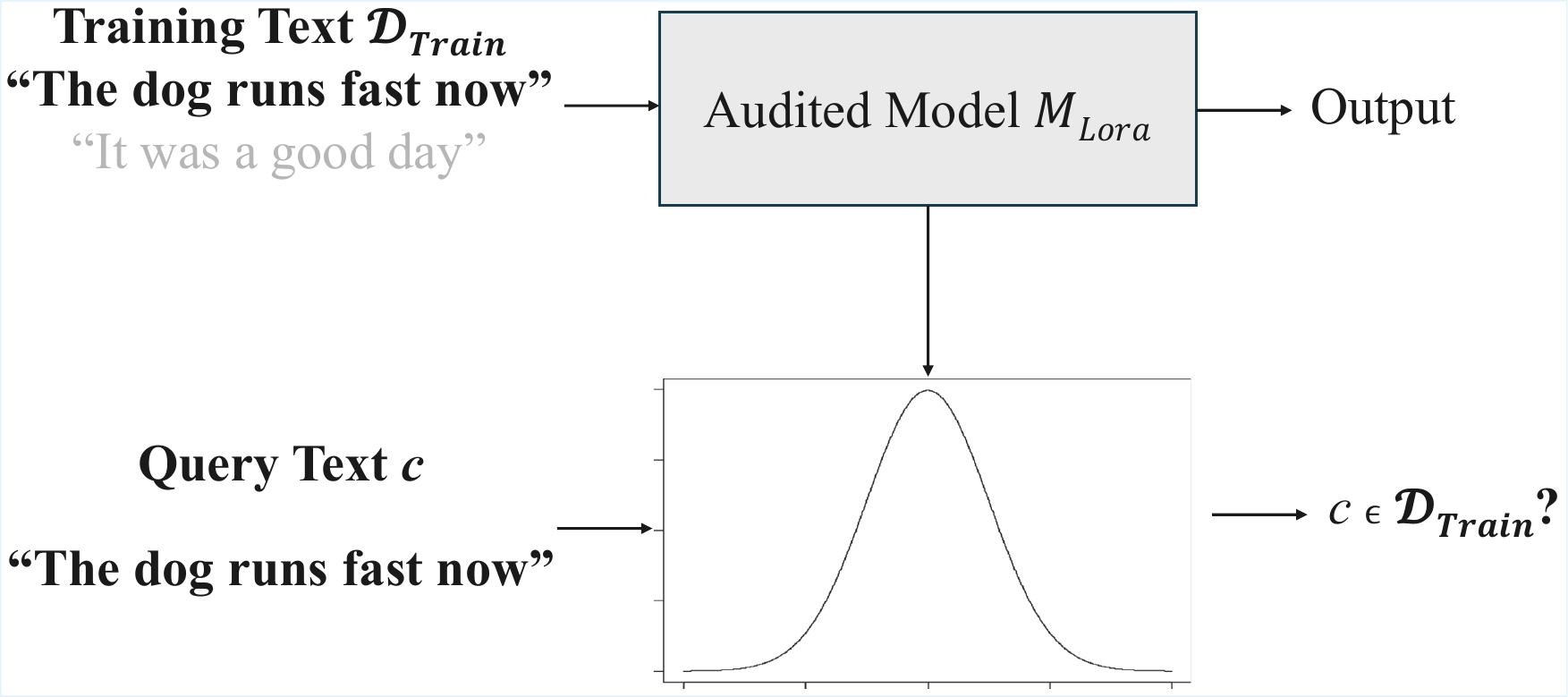}
    \caption{
    The objective of LoRA-MINT is to determine whether a given sample $c$ belongs to the training set of a LoRA-fine-tuned model $\mathcal{D}_{\text{train}}$ by comparing it against a distribution of synthetic samples. 
    }
    \label{fig:concept}
\end{figure}

Along this like of work, the present paper presents LoRA-MINT, a framework designed to assess training data exposure in parameter-efficient fine-tuning. Our main contributions:

\begin{itemize}
    \item We introduce LoRA-MINT, a framework for evaluating if given data were used in LoRA-based fine-tuning.
    \item We perform a comprehensive evaluation of four LLMs and three datasets, analyzing perplexity, distributions, and membership inference accuracy.
    \item The experiments show that LoRA-MINT effectively distinguishes between training and non-training samples, resulting in a useful training data auditing tool for fine-tuned LLMs (see Fig.~\ref{fig:concept}).
\end{itemize}

For the sake of concreteness and current relevance, our discussion and experiments are centered on LoRA-adjusted LLMs, but note that most of the presented methodology is easily applicable for auditing training data given any other technique for adapting LLMs or, more generally, any other domain-adapted AI models.

\section{Related Works}
\subsection{Membership Inference Test (MINT)}

MINT (Membership Inference Test) is a framework developed to evaluate the compliance of ML models with emerging regulatory requirements regarding the use of training data. Its goal is to provide a systematic and robust methodology that can be applied across different data modalities—including text, images, audio and multimodal models—while accommodating various levels of transparency regarding the model’s development and training pipeline \cite{dealcala2024comprehensive,dealcala2024my,dealcala2025gmint,mancera2025my}. Depending on the context, auditors may have limited information about the model or detailed access to its architecture and training procedures, making MINT adaptable to a wide spectrum of regulatory scenarios \cite{dealcala2025mintdemo,dealcala2026mintdemo2}.

The conceptual foundation of MINT is rooted in the extensive body of work on Membership Inference Attacks (MIA), which investigate whether a model has memorized specific training examples \cite{shokri2017membership}. Early contributions demonstrated that ML models can reveal membership information through their output distributions or confidence scores \cite{carlini2022membership}. Subsequent research expanded these ideas, analyzing the role of overfitting, loss-based indicators, and gradient signals as potential membership signals.

In the context of LLMs, MIAs encounter distinctive difficulties arising from the immense scale and heterogeneity of training data, as well as the strong overlap between training and non-training text distributions \cite{yao2024survey}. The single-epoch nature of pre-training further limits the attacker’s ability to observe repeated exposure to samples, making membership detection inherently complex \cite{duan2024membership}. Several techniques have been proposed to address these challenges. For example, \cite{mattern2023membership} employed loss-based comparisons using semantically similar sentences generated through masked LLMs, while \cite{kandpal2023user} introduced user inference strategies that rely on estimating the likelihood of user-specific samples. Recently, new attacks that specifically exploit the probabilistic and structural characteristics of LLMs have been introduced: Min-K \% \cite{shi2023detecting} and Min-K \% ++ \cite{zhang2024min} leverage differences in token probability distributions, while MoPe focuses on the local smoothness of model behavior around training points \cite{li2023mope}. However, the trustworthiness of some of these approaches is limited by inadequate evaluation methodologies that may overestimate their effectiveness \cite{meeus2025sok}. In addition, the intrinsic variability and ambiguity of natural language obscure consistent behavioral signals, making it difficult for attackers to extract robust membership cues \cite{mireshghallah2022quantifying}.

\subsection{Low-Rank Adaptation (LoRA)}
Low-Rank Adaptation (LoRA)~\cite{hu2022lora} is a parameter-efficient fine-tuning (PEFT) technique designed to adapt large pre-trained LLMs to downstream tasks without updating all model parameters. 
Instead of modifying the full weight matrices during fine-tuning, LoRA introduces a pair of low-rank matrices that approximate the weight updates within a lower-dimensional subspace. 
Formally, given a weight matrix $W_0 \in \mathbb{R}^{d \times k}$, LoRA freezes $W_0$ and learns two smaller matrices $A \in \mathbb{R}^{d \times r}$ and $B \in \mathbb{R}^{r \times k}$, where $r \ll \min(d, k)$. 
The adapted weight is expressed as:

\begin{equation}
W = W_0 + BA
\label{eq:W}
\end{equation}

This low-rank decomposition significantly reduces the number of trainable parameters —often by several orders of magnitude— while maintaining competitive downstream performance. 
By updating only the LoRA parameters during training, the original model remains intact, allowing efficient storage, modular fine-tuning, and task switching. 

Beyond efficiency, LoRA also modifies the optimization dynamics of fine-tuning, potentially influencing how the model memorizes and represents information. 
This characteristic makes LoRA-based models an interesting target for MINT and privacy-oriented audits, as the separation between frozen and trainable parameters may affect the distribution of private information within the model.

\section{Proposed Method: LoRA-MINT}

LoRA-MINT is a novel framework designed to identify training samples embedded within LLMs that have been fine-tuned using LoRA. The core principle of the method relies on \textit{perplexity}, a standard measure of how well a language model predicts a given sequence. Perplexity quantifies the model's uncertainty: lower perplexity indicates that the model assigns higher probability to the observed text, suggesting that similar samples may have appeared in its training data. Formally, for a sequence $X$ of $N$ tokens with probabilities $p(x_i)$, the perplexity PPL is defined as:

\begin{equation}
\text{PPL}(X) = \exp\left(-\frac{1}{N} \sum_{i=1}^{N} \log p(x_i)\right)
\label{eq:ppl}
\end{equation}

Using the behavior of perplexity under LoRA-fine-tuned weight adaptations, LoRA-MINT systematically compares perplexity patterns across candidate samples to detect those likely used during fine-tuning. This enables rigorous auditing of data provenance and improves transparency in parameter-efficient adaptation workflows.

In practice, LoRA-MINT leverages \emph{synthetic in-domain test samples} to create a reference distribution of perplexities for text that is known not to be part of the fine-tuning data. Candidate samples are then compared to this reference, with samples exhibiting perplexities below the mean of the synthetic distribution flagged as likely training members, while those above are considered non-members. This approach provides a simple, principled, and efficient method for auditing the use of training data in fine-tuned LLMs. A more detailed explanation of the method, including the construction and refinement of the reference distribution, is provided in Section~\ref{sec:problem-formulation}.

\subsection{Problem Formulation}
\label{sec:problem-formulation}

Let $\mathcal{M}_{\text{LoRA}}$ denote a LoRA-fine-tuned LLM, and let $\mathcal{C} = \{c_i\}_{i=1}^{N_c}$ be the set of candidate samples to audit. Our goal is to determine, for each candidate $c_i$, whether it belongs to the set of training samples $\mathcal{D}_{\text{train}}$ or to the set of non-member samples $\mathcal{D}_{\text{non}}$. To achieve this, we leverage the concept of \emph{perplexity}, as defined in Equation~\eqref{eq:ppl}, which measures how well a model predicts a given sequence of tokens.  

To establish a reference for non-member samples, we generate a set of $M$ \emph{synthetic in-domain samples} $\mathcal{S} = \{s_j\}_{j=1}^{M}$, which are topically aligned with the fine-tuning domain and constructed so as to avoid inclusion in $\mathcal{D}_{\text{train}}$. Using the set of perplexities $\{\mathrm{PPL}(s_j)\}_{j=1}^{M}$ of the synthetic samples, we denote the empirical mean and standard deviation of the reference distribution as $\mu$ and $\sigma$.  

To reduce the influence of extreme values that are far from typical in-domain behavior and better approximate the target distribution, we define a filtered synthetic set $\mathcal{S}'$ by excluding samples outside the percentile range defined by $\theta_{\text{low}}$ and $\theta_{\text{high}}$: 

\begin{equation}
\begin{split}
\mathcal{S}' = \{\, s_j \in \mathcal{S} \mid 
\mathrm{PPL}(s_j) \in 
[\mathrm{PPL}_{\theta_{\text{low}}}, 
\mathrm{PPL}_{\theta_{\text{high}}}] \,\}.
\end{split}
\label{eq:filtered-s}
\end{equation}

\begin{figure}[t]
    \centering
    \includegraphics[width=0.85\linewidth]{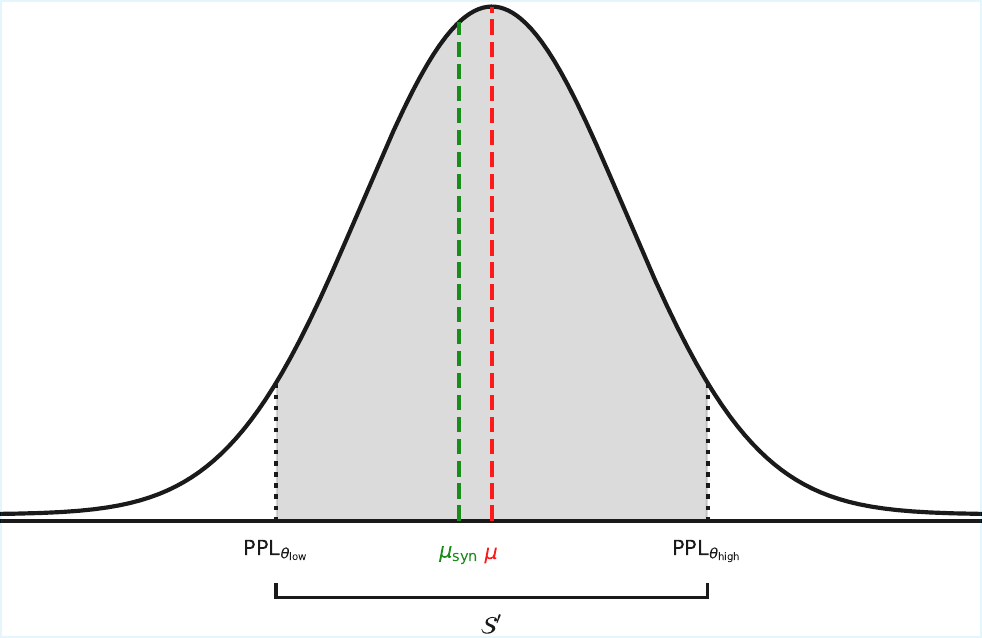}
    \caption{
    Filtered distribution $\mathcal{S}'$ of synthetic in-domain samples, obtained by removing extremes below $\theta_{\text{low}}$ and above $\theta_{\text{high}}$. 
    }
    \label{fig:filtered-distribution}
\end{figure}

The reference distribution is then refined by adjusting its mean as $\mu_{\text{syn}} \leftarrow \mu - \sigma / k$. The values of $\theta_{\text{low}}$, $\theta_{\text{high}}$, and $k$ are jointly selected based on a subset of samples from $\mathcal{D}_{\text{train}}$, choosing the configuration that maximizes detection performance over both member and non-member samples. This procedure allows the reference distribution to better approximate the typical perplexities of $\mathcal{D}_{\text{train}}$ (see Fig.~\ref{fig:filtered-distribution}). This step is crucial because the most challenging cases arise when a sample from $\mathcal{D}_{\text{non}}$ is very similar to a sample from $\mathcal{D}_{\text{train}}$.  

For each candidate sample $c_i \in \mathcal{C}$, we compute its perplexity $\mathrm{PPL}(c_i)$ as defined in Eq.~\eqref{eq:ppl} and compare it with the refined synthetic reference distribution. Candidate samples are classified as:

\begin{equation}
c_i \in 
\begin{cases}
\mathcal{D}_{\text{train}}, & \text{if } \mathrm{PPL}(c_i) < \mu_{\text{syn}}, \\
\mathcal{D}_{\text{non}}, & \text{if } \mathrm{PPL}(c_i) \ge \mu_{\text{syn}}.
\end{cases}
\label{eq:classification}
\end{equation}

\begin{figure*}[t]
\centering
\includegraphics[width=\textwidth]{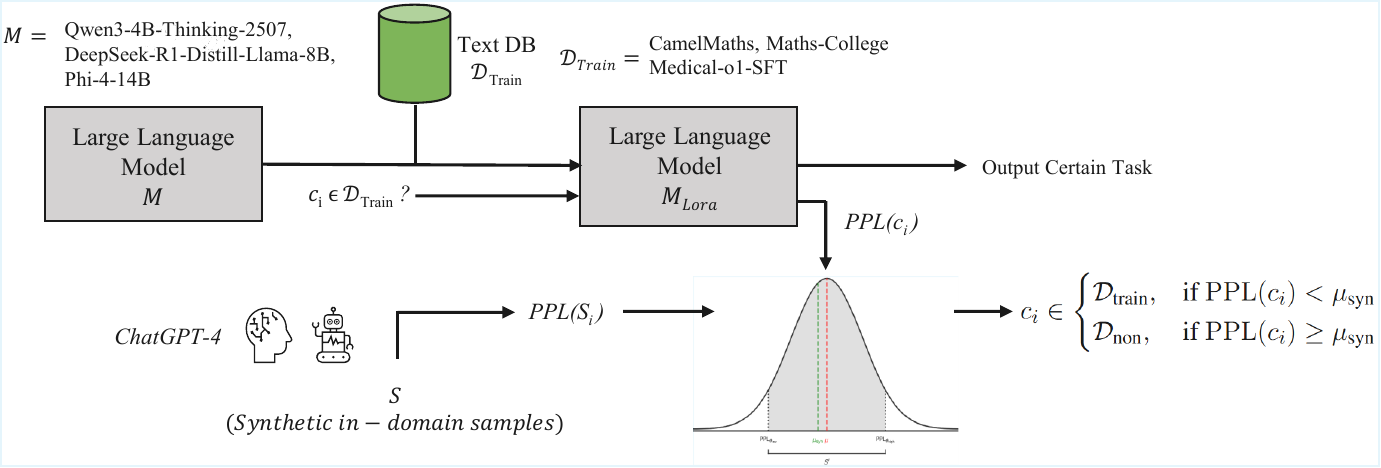}
\caption{Overview of LoRA-MINT. The base LLM is fine-tuned with LoRA using the training set $\mathcal{D}_{\text{train}}$. Synthetic in-domain samples are used to build a reference perplexity distribution, against which the perplexity of each candidate sample $c_i$ is compared to determine whether it belongs to the training set.}
\label{fig:esquemamint_ext}
\end{figure*}

This formulation provides a simple and statistically grounded approach to auditing LoRA-fine-tuned models. By carefully filtering extreme percentiles $(\theta_{\text{low}}, \theta_{\text{high}})$ in $\mathcal{S}$ and adjusting the mean of the reference distribution to be close to $\mathcal{D}_{\text{train}}$, the method accurately captures typical behavior in the domain while remaining robust, even when samples from $\mathcal{D}_{\text{non}}$ are very similar to those from $\mathcal{D}_{\text{train}}$. The method can be summarized as a pipeline (see Fig.~\ref{fig:esquemamint_ext}): 
(i) generate synthetic in-domain samples $\mathcal{S}$, 
(ii) determine and exclude extreme percentiles $(\theta_{\text{low}}, \theta_{\text{high}})$ to refine the reference distribution, 
(iii) adjust the mean of the synthetic distribution by a fraction of the standard deviation to approximate $\mathcal{D}_{\text{train}}$, 
(iv) compute perplexities $\mathrm{PPL}(c_i)$ for candidate samples $\mathcal{C}$, and 
(v) classify candidates relative to the reference mean.

Note that the same spirit of using distribution tails as the core of our method is also found in related work around bias evaluation in biometrics \cite{Imanol26cei} and general AI systems \cite{2023_COMPSAC_BiasAI_N-sigma_DeAlcala}.

\section{Experimental Framework}


\subsection{Audited Large Language Models}\label{audited}
To investigate whether variations in model architecture, size, and complexity influence the effectiveness of MINT, we evaluated four different pre-trained LLMs, namely: Qwen3-4B-Thinking-2507 \cite{yang2025qwen3}, DeepSeek-R1-Distill-Llama-8B \cite{guo2025deepseek}, Phi-4-14B \cite{abdin2024phi}, Llama-3.2-3B, \textbf{Mistral} and \textbf{MistralSmall}. These models were specifically selected to represent a diverse set of LLMs, ranging from smaller optimized architectures to larger, more complex transformers. By auditing models with such varied characteristics, we aim to gain a deeper understanding of how differences in model design affect the training data exposure and auditability. 


\subsection{Experimental Protocol} \label{experimentalprotocaol}
In this work, all LoRA-fine-tuned models $\mathcal{M}_{\text{LoRA}}$ described in Section~\ref{audited} were trained on three different datasets —CAMELMaths Instruction Dataset, Maths-College, and Medical-o1-SFT— following a consistent and carefully controlled parameter-efficient adaptation pipeline. This unified setup ensures that any observed differences in auditing performance can be attributed to the characteristics of the models or datasets, rather than to variations in hyperparameters or training procedures.

To perform the fine-tuning, we adopt LoRA with a fixed scaling parameter $\alpha = 32$, and apply a dropout rate of $0.15$ to regularize the low-rank updates introduced during training. LoRA adapters are inserted into seven key transformer's architecture components, including the attention modules \texttt{q\_proj}, \texttt{k\_proj}, \texttt{v\_proj}, \texttt{o\_proj} and the MLP modules \texttt{gate\_proj}, \texttt{up\_proj}, \texttt{down\_proj}, thus modifying both the attention mechanisms and the feed-forward networks. This placement enables sufficiently expressive parameter-efficient adaptation while leaving the original pretrained model weights untouched.

\begin{table*}[t]
\centering
\caption{
Results of LoRA-MINT. All LoRA-fine-tuned models were trained for a total of 8 epochs, and the auditing procedure was conducted using the proposed synthetic-reference method. 
For each dataset, synthetic distributions were refined by removing lower and upper percentiles, 
$\mathcal{P}_{\text{low}}$ and $\mathcal{P}_{\text{high}}$, respectively, followed by mean adjustment using a scaling factor $k$. 
The reported configuration corresponds to the values of $\mathcal{P}_{\text{low}}$, $\mathcal{P}_{\text{high}}$, and $k$ that maximize detection performance. 
The table reports the AUC and the TPR\,|\,FPR metrics, showing the detection effectiveness across models and datasets.
}
\label{table1}
\resizebox{\textwidth}{!}{%
\begin{tabular}{lccc c ccc}
\hline
 & \multicolumn{3}{c}{\textbf{AUC}} &  & \multicolumn{3}{c}{\textbf{TPR | FPR}} \\ \cline{2-4} \cline{6-8} 
 & CamelMaths & Maths-College & Medical-o1-SFT &  & CamelMaths & Maths-College & Medical-o1-SFT \\ \hline
\multicolumn{1}{l|}{Qwen3-4B}       & 0.791 & 0.890 & 0.841 &  & 0.7325 | 0.121 & 0.804 | 0.024 & 0.740 | 0.112 \\
\multicolumn{1}{l|}{DeepSeek-R1}    & 0.774 & 0.857 & 0.832 &  & 0.722 | 0.224  & 0.785 | 0.175 & 0.764 | 0.140 \\
\multicolumn{1}{l|}{Phi-4}          & 0.782 & 0.900 & 0.842 &  & 0.745 | 0.180  & 0.832 | 0.032 & 0.792 | 0.108 \\
\multicolumn{1}{l|}{Llama-3.2-3B}   & 0.770 & 0.820 & 0.874 &  & 0.780 | 0.240  & 0.790 | 0.150 & 0.800 | 0.052 \\
\multicolumn{1}{l|}{Mistral}        & 0.810 & 0.810 & 0.830 &  & 0.765 | 0.145  & 0.805 | 0.180 & 0.830 | 0.111 \\
\multicolumn{1}{l|}{MistralSmall}  & 0.886 & 0.925 & 0.890 &  & 0.860 | 0.015  & 0.842 | 0.028 & 0.825 | 0.045 \\ \hline
\end{tabular}
}
\end{table*}

All models were fine-tuned for a total of 8 epochs using 15{,}000 task-specific samples per dataset. This protocol therefore defines the precise and reproducible configuration used to obtain the LoRA-finetuned models later subjected to auditing.

To construct the reference distribution required for LoRA-MINT, we generated 1,000 synthetic in-domain samples using GPT-4 together with a subset of 500 samples from $\mathcal{D}_{\text{train}}$, which was used to determine the optimal values of $\theta_{\text{low}}$, $\theta_{\text{high}}$ and $k$, defining the filtered distribution $\mathcal{S}'$ that maximizes detection performance. These synthetic sequences were curated to match the topic and structural properties of the target domain, while being guaranteed not to overlap with the original fine-tuning data. As a result, the resulting reference distribution $\mathcal{S}'$ provides a stable and controlled approximation of the model’s behavior in non-member samples, which is crucial for establishing the baseline perplexity distribution against which membership candidates are evaluated.

After constructing the synthetic reference distribution, we sampled 2,000 training instances and 2,000 held-out test instances from each dataset. For every sample, we computed the perplexity under $\mathcal{M}_{\text{LoRA}}$ and compared it against the refined synthetic distribution following the methodology introduced in Section~3. This procedure enables a systematic and rigorous MINT evaluation, allowing us to quantify the extent to which LoRA-based parameter-efficient fine-tuning introduces measurable privacy leakage across different models and domains.

\section{Experiments and Results}
The experiments were conducted on the models described in Section~\ref{audited}. Each model was fine-tuned using LoRA under the same configuration reported in Section~\ref{experimentalprotocaol}, ensuring a consistent and fully reproducible experimental setup.

For the fine-tuning stage, each model was trained on 15{,}000 samples from the training dataset, and we generated a synthetic reference distribution of 1{,}000 in-domain samples, which serves as a baseline for modeling the expected behavior of non-member instances.

During the evaluation phase, we classified 2{,}000 test samples and 2{,}000 training samples following the auditing procedure described in Section \ref{experimentalprotocaol}. The resulting metrics for each model and dataset are summarized in Table~\ref{table1}.

Importantly, the selection of the percentile thresholds $P_{\text{low}}$, $P_{\text{high}}$, together with the adjustment factor $k$, was guided by performance optimization. For each configuration, the reported values correspond to those that yield the best results. This is evident compared to the unadjusted case, where $P_{\text{low}} = 0$, $P_{\text{high}} = 100$, and $k = 1$ lead to less discriminative AUC values. Consequently, the optimized thresholds provide a more robust choice for standardizing the LoRA-MINT procedure in practical auditing scenarios.

\subsection{Ablation Study of LoRA Target Modules in LoRA-MINT}\label{ablation}

To quantify the importance of different injection points of LoRA within the transformer architecture, we conduct a structured ablation study with five configurations using the \texttt{DeepSeek-R1-Distill-Llama-8B} model on the \texttt{Medical-o1-SF} dataset. This setup allows us to evaluate how attention- and MLP-based adaptations affect detection performance, perplexity, and the controlled overfitting behavior of LoRA-MINT. The results are summarized in Table~\ref{tab:lora_ablation}.

We first evaluate partial adaptation by enabling only \texttt{q\_proj} and \texttt{k\_proj}, followed by the full adaptation of attention with \texttt{q\_proj}, \texttt{k\_proj}, \texttt{v\_proj} and \texttt{o\_proj}. We then isolate the MLP effects using only \texttt{up\_proj} and \texttt{down\_proj}, and subsequently enable the entire MLP pathway, including \texttt{gate\_proj}. Finally, we activate all modules simultaneously, representing the maximum LoRA capacity.

These five experiments provide a decomposition of how different structural components contribute to LoRA-MINT. By isolating partial and complete attention pathways, partial and complete MLP transformations, and their joint activation, this ablation study reveals which mechanisms are most responsible for improving detection performance and shaping the beneficial overfitting dynamics observed during training.

\begin{table}[t]
\centering
\caption{
Ablation study of LoRA target modules in LoRA-MINT using the \texttt{DeepSeek-R1-Distill-Llama-8B} model evaluated on the \texttt{Medical-o1-SF} dataset. Active LoRA injection points are shown in bold, while inactive modules appear in gray. Each configuration isolates a specific subset of attention or MLP pathways to quantify their contribution to detection performance.
}\label{tab:lora_ablation}

\begin{tabular}{lll}
\hline
\multicolumn{1}{c}{Target Modules} & & \multicolumn{1}{c}{AUC} \\
\hline

\begin{tabular}[c]{@{}l@{}}
\textbf{q\_proj}, \textbf{k\_proj}, 
\textcolor{gray}{v\_proj}, \textcolor{gray}{o\_proj},\\
\textcolor{gray}{gate\_proj}, \textcolor{gray}{up\_proj}, \textcolor{gray}{down\_proj}
\end{tabular}
& & 0.504 \\
\hline

\begin{tabular}[c]{@{}l@{}}
\textbf{q\_proj}, \textbf{k\_proj}, \textbf{v\_proj}, \textbf{o\_proj},\\
\textcolor{gray}{gate\_proj}, \textcolor{gray}{up\_proj}, \textcolor{gray}{down\_proj}
\end{tabular}
& & 0.568 \\
\hline

\begin{tabular}[c]{@{}l@{}}
\textcolor{gray}{q\_proj}, \textcolor{gray}{k\_proj}, \textcolor{gray}{v\_proj}, \textcolor{gray}{o\_proj},\\
\textcolor{gray}{gate\_proj}, \textbf{up\_proj}, \textbf{down\_proj}
\end{tabular}
& & 0.688 \\
\hline

\begin{tabular}[c]{@{}l@{}}
\textcolor{gray}{q\_proj}, \textcolor{gray}{k\_proj}, \textcolor{gray}{v\_proj}, \textcolor{gray}{o\_proj},\\
\textbf{gate\_proj}, \textbf{up\_proj}, \textbf{down\_proj}
\end{tabular}
& & 0.754 \\
\hline

\begin{tabular}[c]{@{}l@{}}
\textbf{q\_proj}, \textbf{k\_proj}, \textbf{v\_proj}, \textbf{o\_proj},\\
\textbf{gate\_proj}, \textbf{up\_proj}, \textbf{down\_proj}
\end{tabular}
& & 0.832 \\
\hline

\end{tabular}

\end{table}

\section{Conclusions}

We introduced LoRA-MINT, a lightweight training data auditing framework for LoRA-finetuned LLMs. The method leverages perplexity distributions from synthetic in-domain samples to distinguish training from non-training data in a simple and interpretable way. For the sake of concreteness and current relevance, our discussion and experiments are centered on LoRA-adjusted LLMs, but note that most of the presented methodology is easily applicable for auditing training data given any other technique for adapting LLMs or, more generally, any other domain-adapted AI models.

Across three model families and domain-specific datasets, LoRA-MINT achieves excellent performance, with AUC values ranging from 0.77 to 0.90, showing that detecting training data is possible even when only a small fraction of parameters are fine-tuned. Our analysis shows that refining the synthetic distribution through percentile filtering and mean adjustment is key to maximizing separability, while controlled overfitting increases the gap between member and non-member perplexities. Finally, an ablation study confirms that both attention and MLP adaptations contribute to detection, with their joint activation yielding the best results. In general, LoRA-MINT provides an interpretable and model-agnostic tool for training data auditing in parameter-efficient fine-tuned models.

Future work includes extending the framework to other PEFT methods and developing mitigation strategies to reduce training data memorization while preserving utility \cite{2021_TPAMI_SensitiveNets_Morales}.

\section{Acknowledgment}

Support by M2RAI (PID2024-160053OB-I00, MICIU/FEDER) and Cátedra ENIA UAM-VERIDAS en IA Responsable (NextGenerationEU PRTR TSI-100927-2023-2). Mancera is supported by FPI-PRE2022-104499 MICINN/FEDER and DeAlcala is supported by FPU21/05785 MIU. Work conducted within the ELLIS Unit Madrid.

\bibliographystyle{IEEEtran}
\bibliography{IEEEexample}

\end{document}